\documentclass{article}
\usepackage{spconf,amsmath,graphicx}
\usepackage{multirow}
\usepackage{enumitem}
\usepackage{hyperref}
\newcommand{\githublink}{\href{https://github.com/SebastianEndtTHI/t1t2-detr}{github.com/SebastianEndtTHI/t1t2-detr}}
\setlist{nosep, leftmargin=14pt}

\title{Detection Instead of Inversion: T1-T2 Microstructure Quantification With Detection Transformers}
\name{Sebastian Endt $^{1,2}$, Fatih Özkan $^{1}$, Marion I. Menzel$^{1,3}$}
\address{$^1$ AImotion Bavaria, Technische Hochschule Ingolstadt, Ingolstadt, Germany\\$^2$ TUM School of Computation, Information and Technology, Technical University of Munich, Germany\\$^3$ TUM School of Natural Sciences, Technical University of Munich, Germany}
\begin{document}
%
\maketitle
%
%
\begin{abstract}
T1-T2 correlation MRI has strong potential to quantify microstructural compartments without imposing a biological model. Traditionally, full T1-T2 spectra per voxel are reconstructed with an ill-conditioned and time-consuming Inverse Laplace Transform. We propose to view this challenge as an object detection task: A Detection Transformer (DETR) takes 64 TI/TE signals and directly predicts a set of compartments specified by T1, T2, and signal fraction. We achieve a mean average precision mAP@10 of 76.8\% on simulated data, outperforming both our previous model design (53.8\%) and regularized NNLS (66.1\%). We show that our DETR approach outperforms NNLS in almost every setting across compartment count, compartmental signal fraction, and SNR.
Reconstruction is orders of magnitude faster than NNLS.

The results are fully reproducible from our public code: \githublink
\end{abstract}
\begin{keywords}
Magnetic Resonance Imaging, T1-T2 correlation imaging, multicomponent relaxometry, detection transformer, set prediction, inverse Laplace transform
\end{keywords}
%
%
\section{Introduction}\label{sec:intro}
\begin{figure*}[!t]
    \centering
    \includegraphics[width=\textwidth]{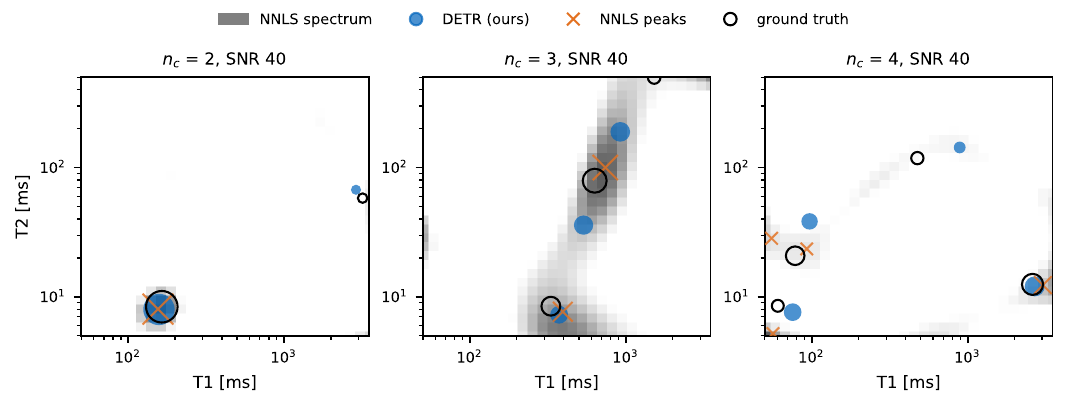}
    \caption{T1-T2 spectra of random test voxels with $n_c\in\{2,3,4\}$ at SNR\,=\,40. Grayscale: NNLS spectrum; orange crosses: NNLS compartments; filled blue circles: DETR (ours); empty black circles: ground truth. Marker size represents signal fraction.}
    \label{fig:spectra}
\end{figure*}

Magnetic Resonance Imaging (MRI) voxels contain a mixture of microstructural compartments, such as intra- and extracellular water, myelin water, and free water -- each with distinct relaxation properties. Multidimensional correlation MRI, e.g., T1-T2 correlation imaging, can distinguish these compartments \cite{english1991,Benjamini2020review,Slator2021review}. Conventionally, full multidimensional parameter spectra are reconstructed for each voxel. This ill-posed Inverse Laplace Transform (ILT) is usually solved using regularized NNLS algorithms.

Earlier work on diffusion correlation MRI showed that the quantification of microstructural compartments can be interpreted as an object detection task \cite{endt2026}. A detection transformer (DETR) \cite{carion2020} is used for end-to-end prediction of a set of compartments. The permutation invariance in this set prediction problem is solved with Hungarian matching \cite{hungarian}.



\noindent In this work we
\begin{itemize}
    \item transfer the diffusion MRI DETR \cite{endt2026} to T1-T2 correlation MRI, with log-scaled T1/T2 targets, uniform compartment weighting, and other refinements.
    \item systematically evaluate DETR against NNLS across compartment count, signal fraction, and SNR.
\end{itemize}
%
\section{Methods}\label{sec:methods}
\subsection{Signal simulation}
We simulate inversion-recovery spin-echo data with varying inversion time TI and echo time TE, i.e., a T1-T2 correlation imaging protocol sensitive to myelin water \cite{endt2023}. The protocol covers a total of 64 contrasts. The signal of a voxel with $n_c$ compartments is given as:
\begin{equation}
    \mathbf{S} = \sum_{k=1}^{n_c} f_k \left(1 - 2e^{-\mathbf{TI}/T_{1,k}} + e^{-\mathrm{TR}/T_{1,k}}\right) e^{-\mathbf{TE}/T_{2,k}},
\end{equation}
where $f_k$ is the compartment's signal fraction, or weight. The data set comprises $n_c$ between 1 and 4 with equal shares. All $f_k$ in a voxel contribute at least $5\%$ and sum up to 1. Parameters are drawn from log-uniform distributions $T_1\in[50, 3500]\,\mathrm{ms}$, $T_2\in[5, 500]\,\mathrm{ms}$, with $T_1>T_2$.

Gaussian noise with random $\mathrm{SNR}\in[10,150]$ is added as $\sigma=\max|\mathbf{S}|/\mathrm{SNR}$, followed by normalization to $\|\mathbf{S}\|_2=1$.

Training, validation, and test data contain 200k, 25k, and 25k random voxels, respectively. Additionally, 5k voxels with SNR $ \in \{5, 10, 20, 40, 80, 120, 200, \infty\}$, using one fixed noise pattern, yield 8 data sets differing only in noise amplitude to investigate SNR dependency.

\subsection{Detection Transformer}
The detection transformer (DETR) \cite{carion2020} is adapted from an earlier work on diffusion correlation MRI \cite{endt2026}. An MLP encoder compresses the 64 input signals into a hidden representation. A transformer decoder takes ten learned queries and has them cross-attend to these signals. Separate sigmoid-activated MLP prediction heads predict T1, T2, and signal fraction for each query. Existence scores for all queries are predicted jointly, taking all other queries into account. The model has $\approx4.5\times10^6$ learnable parameters.

Given an unordered set of predicted compartments, the assignment to ground truths is handled with Hungarian matching \cite{hungarian}. Then the loss is computed from mean squared errors of T1, T2 and signal fraction, and the binary cross entropy of the existence score. Existence loss is weighted with 0.1 compared to the other loss terms. AdamW optimizer learning rate starts at $10^{-4}$ and is halved after 7 epochs without improvement. Training is stopped after 35 epochs without improvement; the checkpoint with the best validation mAP@10 is saved. The metric mAP@10 is the mean average precision, where a prediction matches if its deviations in log(T1) and log(T2) each lie within $\tau=10\%$ of the log parameter range \cite{pascalvoc,mscoco,endt2026}. The existence score threshold is chosen such that F1-score is maximized on validation data.

In our diffusion correlation proof-of-concept \cite{endt2026}, parameters were scaled linearly, which is problematic for relaxation times, and both training loss and matching cost were weighted by signal fraction, underrepresenting small compartments. Therefore we compare two models:
\begin{itemize}
    \item DETR (diff): linear parameter scaling $T_{1/2}/T_{1/2,\mathrm{max}}$; Loss/cost weighted with signal fraction.
    \item DETR (ours): logarithmic parameter scaling $(\log T_{1/2}-\log T_{1/2,\mathrm{min}})/(\log T_{1/2,\mathrm{max}}-\log T_{1/2,\mathrm{min}})$;  Equal weighting of compartments in loss/cost.
\end{itemize}
Otherwise, both variants use the same architecture, training and data, including other refinements over \cite{endt2026}.

\subsection{NNLS baseline}
\begin{table*}[!t]
    \centering
    \caption{Quantitative results on 25,000 voxel test data. mAP@10 is additionally broken down for different numbers of compartments $n_c$ and compartmental signal fractions $f$.}
    \label{tab:results}
    \begin{tabular}{l|ccc|cccc|ccc}
        \hline
        \multirow{2}{*}{Method} & mAP@5 & mAP@10 & Strict & \multicolumn{4}{c|}{$n_c$} & \multicolumn{3}{c}{$f$} \\
                    & [\%] & [\%] & acc.\ [\%] & 1 & 2 & 3 & 4 & $<0.35$ & $0.35$--$0.65$ & $\geq 0.65$\\
        \hline
        NNLS        & 43.3 & 66.1 & 40.5 & \textbf{99.0} & 78.2 & 62.1 & 51.4 & 42.8 & 82.6 & 93.3 \\
        DETR (diff) & 25.6 & 53.8 & 39.4 & 98.5 & 74.0 & 50.1 & 39.2 & 33.2 & 73.2 & 93.8 \\
        DETR (ours) & \textbf{49.2} & \textbf{76.8} & \textbf{49.0} & 98.9 & \textbf{89.8} & \textbf{77.4}
                    & \textbf{64.6} & \textbf{60.8} & \textbf{91.9} & \textbf{98.0} \\
        \hline
    \end{tabular}
\end{table*}

\begin{figure*}[t!]
    \centering
    \includegraphics[width=\textwidth]{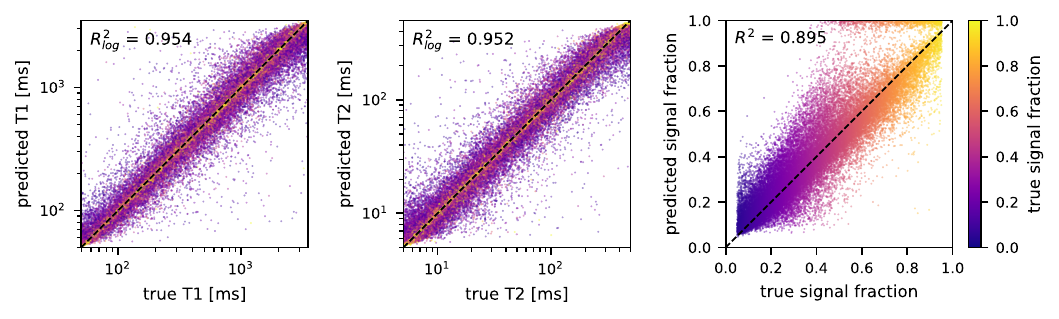}
    \caption{True vs.\ predicted $T_1$ (left), $T_2$ (middle), and signal fraction (right) for all matched DETR (ours) test compartments, colored by true signal fraction.}
    \label{fig:scatter}
\end{figure*}

Our approach is judged against a standard regularized NNLS algorithm. T1-T2 spectra are defined using a $30\times30$ grid within the same parameter ranges. Also obeying $T_1>T_2$, this leaves 771 spectrum entries. Zero-order Tikhonov regularization is applied with respect to the $\chi^2$ criterion such that $\chi^2=1.02\chi^2_\mathrm{min}$. Compartments are identified with hill climbing and surrounding entries are assigned to local maxima. Compartmental T1 and T2 are computed as the assigned entries' weighted geometric mean; compartmental signal fraction as their integral. The peak fraction serves as confidence; its threshold (0.1, best validation F1) only applies to strict accuracy.

\subsection{Evaluation}
All methods are evaluated on previously unseen test data and compared using the same metrics. In addition to mAP@10, we compute mAP@5 at $\tau=5\%$. Strict accuracy denotes the fraction of voxels for which the number of compartments is predicted correctly and all predicted compartments satisfy the $\tau=10\%$ matching criterion. $95\%$ confidence intervals are computed from a paired percentile bootstrap, over the same voxels for each method and using 1000 resamples.


\section{Results}\label{sec:results}
Fig. \ref{fig:spectra} shows random spectra with $n_c\in\{2,3,4\}$, combining ground truth, NNLS spectrum and DETR prediction. Both methods identify large, well-separated compartments. DETR also detects smaller compartments, whereas these are less distinctly resolved by NNLS. NNLS spectra are generally broadened along ridges.

Fig.~\ref{fig:scatter} shows agreement between all matched DETR compartments and the ground truth ($R^2_{\log}=0.954$ for $T_1$ , $0.952$ for $T_2$, $R^2=0.895$ for $f$). Table \ref{tab:results} summarizes the quantitative results on 25,000 previously unseen test voxels. Our proposed DETR achieves a higher mAP@10 of 76.8\% compared to 66.1\% for NNLS and 53.8\% for the diffusion design. At the stricter $\tau=5\%$, the mAP decreases to 49.2\%, 43.3\% and 25.6\%, respectively. Strict accuracy rises from 40.5\% using NNLS to 49.0\% using our DETR.

In single-compartment voxels, all methods show comparably high performance. For multi-compartment voxels, which constitute the clinically interesting cases, our DETR outperforms NNLS by +11.6\,pp (percentage points) for $n_c=2$, +15.3\,pp for $n_c=3$, and +13.3\,pp for $n_c=4$. The largest improvement is observed for small compartments ($f<0.35$), where our DETR achieves an mAP@10 of 60.8\%, compared with 42.8\% for NNLS and 33.2\% for the diffusion design. For medium and large compartments we still outperform the other methods.

\begin{figure}[t]
    \centering
    \includegraphics[width=\linewidth]{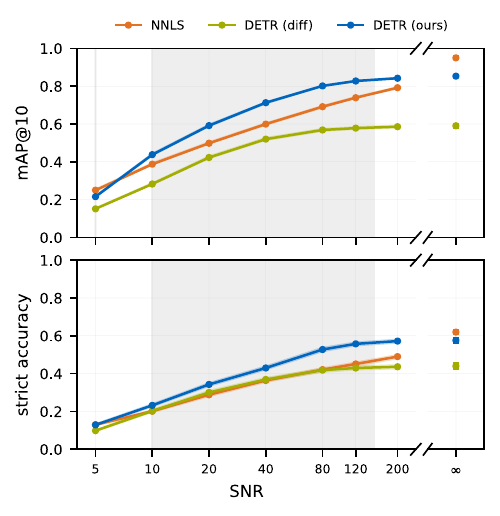}
    \caption{mAP@10 (top) and strict accuracy (bottom) vs.\ SNR on 5000 voxels with identical noise pattern. Shaded: bootstrap $95\%$ CIs; gray: training SNR range.}
    \label{fig:snr}
\end{figure}

Fig. \ref{fig:snr} shows mAP@10 and strict accuracy across the paired SNR data sets. Across SNRs from 10 to 200, our DETR model outperforms NNLS, most prominently for SNR of 40--80 with an increase of +11.3\,pp and +11.0\,pp in mAP@10, respectively. This benefit lies well beyond the confidence intervals. At low SNR of 5, which was not included in the training data, NNLS outperforms DETR. For very high SNR, our model seems to plateau (82.8\% at SNR=120, 85.3\% without noise), while NNLS reaches up to 95.0\% mAP@10 in noise-free data. The diffusion design converges towards 59\%. Strict accuracy yields similar results, with our proposed DETR method outperforming the other methods across SNRs from 10 to 200.

With an inference time of 0.018\,ms per voxel on an NVIDIA A100 (MIG with 40\,GB), DETR is about $8\times10^{3}$ times faster than NNLS (147\,ms per voxel on a single AMD EPYC 7763 CPU core).

\section{Discussion}\label{sec:discussion}
With the introduction of object detection principles to the field of correlation MRI, we propose a potential paradigm shift: While current approaches for spectra reconstruction, like NNLS, reconstruct a full spectrum, we translate the signals directly into a set prediction of compartments.

In voxels with more or smaller compartments, the regularization in NNLS merges similar compartments, and small compartments are lost in broad peaks.

Comparison with our initial design from diffusion correlation MRI shows that the methods do not translate one-to-one from one modality to another. The logarithmic parametrization and equal weighting of compartments in losses improve mAP@10 by 23\,pp. The gain concentrates on small compartments, where the diffusion design falls behind even NNLS. A separate ablation of parametrization and weighting was omitted due to space limitations.

DETR is superior within its training SNR range. In noise-free data, it plateaus while model-based NNLS keeps improving; at SNR\,=\,5, outside the training range, NNLS is also slightly better, although earlier work suggested deep learning to be more robust to noise \cite{endt2021}. Whether a broader training distribution helps is left for future work.

The study is subject to limitations: All data was simulated with Gaussian noise, due to the real-valued signals, and without considering field inhomogeneities, incomplete inversions, exchange, or magnetization transfer. The training follows a specific measurement protocol and would have to be retrained for different acquisition parameters. Finally, the test on in vivo data is the subject of ongoing work.

\section{Conclusion}\label{sec:conclusion}
This study interprets the microstructure quantification with T1-T2 correlation imaging as an object detection task, using a Detection Transformer. Compared to traditional spectra reconstruction with NNLS we reach better results (mAP@10 of 76.8\% vs. 66.1\%) and are orders of magnitude faster.

\section{Compliance with ethical standards}
\label{sec:ethics}
This is a numerical simulation study for which no ethical approval was required. 
\section{Acknowledgments}
\label{sec:ack}
No funding was received for this study. The authors have no relevant interests to disclose. Claude (Anthropic) assisted with code development, text editing and layout. All content was verified by the authors.

\bibliographystyle{IEEEbib}
\bibliography{refs}

\end{document}